\documentclass[11pt, a4paper]{article}

\usepackage[utf8]{inputenc}
\usepackage{geometry}
\usepackage{amsmath, amssymb, amsfonts}
\usepackage{graphicx}
\usepackage[font=small,labelfont=bf]{caption}
\usepackage{hyperref}
\usepackage{xcolor}
\usepackage{booktabs}
\usepackage{authblk}
\usepackage{multirow}
\usepackage[style=numeric, sorting=none, backend=biber]{biblatex}
\usepackage{tikz}
\usetikzlibrary{shapes, arrows.meta, positioning, fit, calc, backgrounds}

\usepackage{tcolorbox}
\tcbuselibrary{skins, breakable}
\newtcolorbox{reprobox}[1][]{
  colback=gray!5!white,
  colframe=gray!75!black,
  title=\textbf{Reproducibility \& Hyperparameter Specification},
  fonttitle=\bfseries\sffamily,
  boxrule=0.8pt,
  arc=2pt,
  #1
}
\newcommand{\R}{\mathbb{R}}

\newcommand{\modelname}{\textsc{SplineGPT}}

\title{\textbf{Kinematic Tokenization: Optimization-Based Continuous-Time Tokens for Learnable Decision Policies in Noisy Time Series}}

\author{Griffin M. Kearney, Ph.D.}
\affil{Syracuse, New York, USA \\ gkearneyengineering@gmail.com}
\date{\today}

\begin{document}

\maketitle

\begin{abstract}
Transformers are designed for discrete tokens, yet many real-world signals are continuous processes observed through noisy sampling. Discrete tokenizations (raw values, patches, finite differences) can be brittle in low signal-to-noise regimes, especially when downstream objectives impose asymmetric penalties that rationally encourage abstention.
We introduce \textit{Kinematic Tokenization}, an optimization-based continuous-time representation that reconstructs an explicit spline from noisy measurements and tokenizes local spline coefficients (position, velocity, acceleration, jerk).
This is applied to financial time series data in the form of asset prices in conjunction with trading volume profiles.
Across a multi-asset daily-equity testbed, we use a risk-averse asymmetric classification objective as a stress test for learnability. Under this objective, several discrete baselines collapse to an absorbing cash policy (the \textit{Liquidation Equilibrium}), whereas the continuous spline tokens sustain calibrated, non-trivial action distributions and stable policies.
These results suggest that explicit continuous-time tokens can improve the \emph{learnability and calibration} of selective decision policies in noisy time series under abstention-inducing losses.
\end{abstract}

\section{Introduction}
The adaptation of Foundation Models, which were originally designed for discrete natural language tokens, to continuous time series data presents a fundamental representation challenge. While text is inherently discrete, physical and economic signals are continuous systems viewed through the lens of a discrete sampling clock. Current approaches often rely on patching or direct quantization of raw values. These methods fail to account for measurement uncertainty or the underlying dynamic laws (i.e. the ``physics'') governing system evolution between samples. While financial markets inherently lack the rigorous first-principle governance found in natural physical systems, in our experiments we impose Newton-like equations of motion on the evolution of trading parameters, utilizing kinematic consistency as a robust inductive bias for representation learning.

\subsection{Physics-Informed Tokenization Framework}
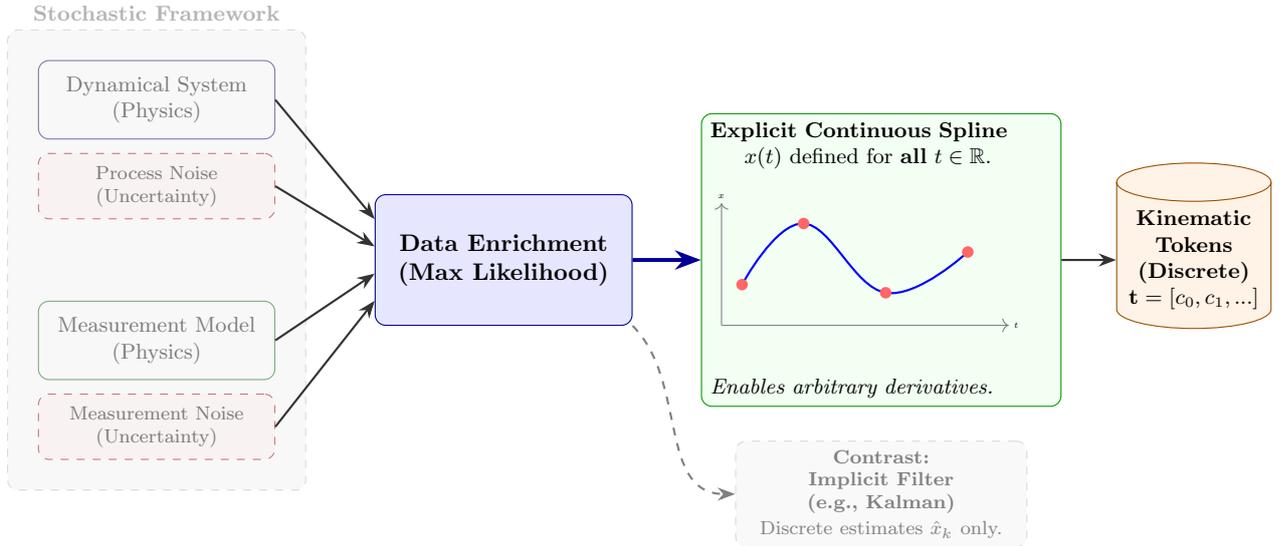
\begin{figure}[htbp]
    \centering
    \begin{tikzpicture}[
        scale=0.9,
        transform shape,
        node distance=1.5cm and 2.0cm,
        auto,
        % --- Styles ---
        conceptNode/.style={
            rectangle, 
            draw=gray!80!black, 
            fill=white, 
            text width=3.2cm, 
            align=center, 
            rounded corners, 
            minimum height=3em,
            font=\footnotesize
        },
        stochastic/.style={
            rectangle, 
            draw=red!60!black, 
            fill=red!5, 
            text width=3.2cm, 
            align=center, 
            rounded corners, 
            minimum height=2.5em,
            font=\scriptsize,
            dashed
        },
        core/.style={
            rectangle, 
            draw=blue!60!black, 
            fill=blue!10, 
            text width=3.5cm, 
            align=center, 
            rounded corners, 
            minimum height=5em,
            font=\small\bfseries
        },
        output/.style={
            rectangle, 
            draw=green!60!black, 
            fill=green!5, 
            text width=5cm, 
            align=left, 
            rounded corners, 
            minimum height=8em, 
            font=\footnotesize
        },
        token/.style={
            cylinder, 
            shape border rotate=90, 
            draw=orange!60!black, 
            fill=orange!10, 
            text width=2.0cm, 
            align=center, 
            minimum height=3.5em,
            aspect=0.25,
            font=\footnotesize\bfseries
        },
        arrow/.style={
            thick, 
            ->, 
            >=Stealth, 
            color=gray!40!black, 
            rounded corners
        }
    ]

        % --- Left Side: Physics & Uncertainty ---
        \node [conceptNode, draw=blue!40!black] (dynamics) {Dynamical System \\ (Physics)};
        \node [stochastic, below=0.2cm of dynamics] (proc_noise) {Process Noise \\ (Uncertainty)};
        
        % Increased spacing here (0.8cm -> 1.2cm) to separate the two groups
        \node [conceptNode, draw=green!40!black, below=1.2cm of proc_noise] (measure) {Measurement Model \\ 
        (Physics)};
        \node [stochastic, below=0.2cm of measure] (meas_noise) {Measurement Noise \\ (Uncertainty)};
        
        % Background Box
        \node[draw=gray!30, fill=gray!10, fill opacity=0.5, rounded corners, dashed, inner sep=0.4cm, fit=(dynamics)(proc_noise)(measure)(meas_noise), label={[gray!60, font=\footnotesize\bfseries]above:Stochastic Framework}] (framework) {};

        % --- Middle: The Engine ---
        \node [core, right=1.0cm of framework] (optimization) {Data Enrichment \\ (Max Likelihood)};
        
        % --- Right Side: The Output Distinction ---
        \node [output, right=1.0cm of optimization] (explicit) {
            \textbf{Explicit Continuous Spline} \\
            \hspace{0.5cm}$x(t)$ defined for \textbf{all} $t \in \R$. \\
            \vspace{3.0cm} 
            \textit{Enables arbitrary derivatives.}
        };
        
        % --- DRAW THE GRAPH ---
        \begin{scope}[shift={(explicit.south west)}, xshift=0.3cm, yshift=1.2cm, scale=0.6]
            \draw[->, gray] (0,0) -- (7,0) node[right, black] {\tiny $t$};
            \draw[->, gray] (0,0) -- (0,3) node[above, black] {\tiny $x$};
            \draw[blue, thick, smooth] plot [tension=0.7] coordinates {(0.5,1.0) (2.0,2.5) (4.0,0.8) (6.0,1.8)};
            \fill[red!60] (0.5,1.0) circle (4pt);
            \fill[red!60] (2.0,2.5) circle (4pt);
            \fill[red!60] (4.0,0.8) circle (4pt);
            \fill[red!60] (6.0,1.8) circle (4pt);
        \end{scope}
        
        % Contrast Box
        \node [output, below=0.5cm of explicit, fill=gray!5, draw=gray!30, dashed, text=gray, minimum height=3em, text width=4cm, align=center] (implicit) {
            \scriptsize \textbf{Contrast: \\Implicit Filter\\ (e.g., Kalman)} \\
            Discrete estimates $\hat{x}_k$ only.
        };

        % --- Far Right: Tokens ---
        \node [token, right=0.8cm of explicit] (tokens) {Kinematic \\ Tokens \\ (Discrete) \\ $\mathbf{t} = [c_0, c_1, ...]$};

        % --- Connections (FIXED CROSSING) ---
        % Top pair connects to upper anchors
        \draw [arrow] (dynamics.east) -- ($(optimization.west)+(0,0.6)$);
        \draw [arrow] (proc_noise.east) -- node[midway, above, pos=0.3, font=\tiny, sloped, text=red!60!black] {} ($(optimization.west)+(0,0.2)$);
        
        % Bottom pair connects to lower anchors (Order fixed: Measure is higher than Meas_Noise)
        \draw [arrow] (measure.east) -- ($(optimization.west)+(0,-0.2)$);
        \draw [arrow] (meas_noise.east) -- node[midway, below, pos=0.3, font=\tiny, sloped, text=red!60!black] {} ($(optimization.west)+(0,-0.6)$);

        \draw [arrow, line width=1.5pt, blue!60!black] (optimization) -- node[midway, above, font=\scriptsize] {} (explicit);
        \draw [arrow, dashed, gray] (optimization.south east) to[out=-45, in=180] (implicit.west);
        \draw [arrow] (explicit) -- node[midway, above, font=\scriptsize] {} (tokens);

    \end{tikzpicture}
    \caption{\textbf{Conceptual Framework for Physics-Informed Continuous Tokenization.} The approach integrates models of system dynamics and measurement protocols, both subject to stochastic uncertainty. Unlike traditional filtering (e.g. Kalman) which yields discrete estimates via implicit representations, our optimization-based approach produces an \textbf{explicit continuous spline} representation. This continuum is necessary to extract high-order kinematic tokens for the foundation model.}
    \label{fig:concept}
\end{figure}

In this paper, we bridge this gap between discrete measurement samples and the continuous underlying processes by situating the time series problem within a framework of \textbf{Physics-Informed AI}. Rather than treating data points as static numbers, we treat them as noisy observations of a latent dynamical system subject to stochastic forcing.
We introduce a methodology based on \textbf{Optimization-Based Data Enrichment} \cite{Kearney2024}. As illustrated in Figure \ref{fig:concept}, our framework fundamentally incorporates physics through two stochastic channels and two coupled manifolds:
\begin{enumerate}
    \item \textbf{Dynamical Systems Model:} A description of how the system state (e.g. Price) and system flux (e.g. Volume) evolve continuously over time, subject to random process noise.
    \item \textbf{Measurement Model:} A description of how we observe these states, distinguishing between \textit{snapshot measurements} (e.g. Price) and \textit{aggregate measurements} (e.g. Volume).
\end{enumerate}

\subsection{Explicit vs. Implicit Representations}
A critical distinction of our approach is the nature of the resulting representation. Traditional methods like Kalman filtering are \textit{implicit}: they produce best estimates at pre-specified discrete filtering times enabling discrete data correction and higher fidelity (but still discrete) sampling. Continuous limits of the Kalman approach do exist \cite{sarkka2023bayesian}, but these treatments fail to produce compact discrete representations (e.g. spline coefficients) which are natural for use in Transformer architectures; these methods are typically implicit and recursive. In contrast, our optimization framework solves for an explicit continuous spline that defines the state $x(t)$ and all its derivatives for \textit{all} times $t$ in the interval via a direct compact representation.

This explicit nature is paramount for tokenization. It allows us to analytically compute high-order derivatives, namely velocity, acceleration, jerk, and volume flow rates, which are often absent in discrete methods but contain vital information about the underlying forces driving the system. These kinematic parameters form the basis of our continuous-time tokens, providing the Transformer with a physically grounded representation of the signal's coupled evolution.

\subsection{Finance as a Stochastic Testbed}
While the experimental validation in this work focuses on daily data in financial markets, the proposed framework is domain-agnostic. Financial time series represent a particularly challenging class of stochastic dynamic systems, characterized by low signal-to-noise ratios and time-varying volatility. However, the underlying optimal spline formulation is generalizable. By recasting the dynamics, process noise and measurement models, this strategy can be adapted to distinct physical domains, such as inertial navigation, robotic control, or other diverse signal processing applications, where preserving the continuous derivatives of the signal is essential.

% -----------------------------------------
% RELATED WORK
% -----------------------------------------
\section{Related Work}

\subsection{Foundation Models for Time Series}
The success of Large Language Models has spurred the development of Large Observation Models (LOMs). Current state-of-the-art architectures like PatchTST \cite{nie2023timeseriesworth64} and iTransformer \cite{liu2024itransformer} utilize patching to capture local semantic information. More recent efforts in 2025, such as Google's TimesFM-ICF \cite{Faw2025InContext}, have introduced in-context learning to adapt to non-stationary data. However, these models remain physically agnostic; they treat time series patches as statistical vectors, ignoring the underlying kinematic continuity of the signal.

\subsection{Emergent Continuous Architectures (SSMs \& LNNs)}
To address the computational inefficiency of Transformers on long sequences, State Space Models (SSMs) like Mamba have gained prominence. Recent applications such as MambaStock \cite{shi2024mambastockselectivestatespace} and TimeMachine \cite{Ahamed2024TimeMachine} leverage Mamba's selective scan mechanism to learn discretized latent dynamics with linear complexity. 
Similarly, Liquid Neural Networks \cite{Hasani2020LTCNN} and Physics-Informed Koopman Networks \cite{liu2025PIKoop} explicitly model neuronal activations or latent states via differential equations to handle irregularity.

While effective, these methods typically operate on raw, noisy discrete inputs and must \textit{learn} the system dynamics (matrix $\mathbf{A}$ or ODE parameters) from scratch. \modelname{} differentiates itself by decoupling the physical extraction from the pattern recognition. We do not ask the model to extract the physics; we explicitly calculate a continuous state (velocity, acceleration, etc) via spline optimization ex-ante, providing the Transformer with pre-computed physically-grounded discrete representations of the continuous physics while retaining the training efficiency of standard attention mechanisms.

\subsection{Physics-Informed Finance}
The application of Physics-Informed Neural Networks (PINNs) \cite{RAISSI2019686} to finance has largely focused on solving known Partial Differential Equations (PDEs). Recent work by Socaciu et al. (2025) \cite{Socaciu2025PINN} demonstrates the utility of PINNs for solving Black-Scholes and Heston equations for option pricing. Deep learning approaches in limit order books typically rely on discrete spatial or temporal structures \cite{Sirignano2019LOB, Zhang_2019}.
However, these approaches are largely deductive—enforcing known laws onto data. Our approach is inductive: we use optimization-based enrichment \cite{Kearney2024} to derive a latent kinematic state from the asset price data stream, enabling alpha generation in trading scenarios where a general governing law is unknown.

It should be noted that in the present approach local ``laws'' are prescribed in the absence of fundamental physics (e.g. stochastically forced second order motion mirroring Newton's second law) which are used for the financial data tokenization. We emphasize that the tokenization scheme that is described here is adaptable to other data domains where rich physical frameworks governing system behavior are proven by modifying the governing laws and measurement models. The same general treatment is applicable leveraging rigorous physics when available to adjust the equivalent data enrichment formulation as we demonstrate in (\ref{eqn:1}) and (\ref{eqn:2}).

\section{Methodology}

\begin{figure}[htbp]
    \centering
    \resizebox{\textwidth}{!}{% Resize to fit page width
    \begin{tikzpicture}[
        node distance=2.5cm and 2.5cm,
        auto,
        block/.style={
            rectangle, 
            draw=blue!60!black, 
            fill=blue!5, 
            text width=4.5cm, 
            align=center, 
            rounded corners, 
            minimum height=3.5em,
            thick
        },
        cloud/.style={
            draw=red!60!black, 
            fill=red!5, 
            dashed, 
            inner sep=0.5cm,
            rounded corners
        },
        arrow/.style={
            thick, 
            ->, 
            >=Stealth, 
            color=gray!40!black
        },
        line/.style={
            thick, 
            color=gray!40!black
        }
    ]

        % --- Nodes ---
        
        % 1. Input
        \node (input) {Raw Time Series};
        
        % 2. Pre-processing
        \node [block, right=of input] (log) {Log Transform \\ $y_k = \ln(P_t)$ \\ $\tilde{y}_k = \ln(V_t)$};
        
        % 3. The Core Contribution (Spline Enrichment)
        \node [block, right=of log, fill=orange!10, draw=orange!60!black, text width=5.5cm] (spline) {
            \textbf{Optimization-Based Data Enrichment} \\ 
            \textit{Max Likelihood Estimation}
        };
        
        % 4. Tokenization
        \node [block, right=of spline] (tokens) {
            Spline Tokens \\ 
            $\mathbf{t}_k = [c_0, c_1, c_2, c_3]^T$ \\
            $\tilde{\mathbf{t}}_k = [\tilde{c}_0, \tilde{c}_1, \tilde{c}_2, \tilde{c}_3,\tilde{c}_4]^T$
        };
        
        % 5. Post-Process (Normalization)
        \node [block, below=of tokens] (norm) {
            Window Anchoring \\ \& Normalization
        };
        
        % 6. Transformer
        \node [block, left=of norm, fill=green!10, draw=green!60!black, minimum width=5.5cm] (transformer) {
            \textbf{\modelname} \\
            \textit{Causal Transformer}
        };
        
        % 7. Output/LoRA
        \node [block, left=of transformer, dashed] (lora) {
            LoRA Adaptation \\ (Downstream Tasks)
        };

        \node [left=of lora] (output) {Forecast / Strategy};

        % --- Connections ---
        
        % Top Row
        \draw [arrow] (input) -- node[midway, above] {\footnotesize Discrete $P_t$} (log);
        \draw [arrow] (log) -- node[midway, above] {\footnotesize Discrete $y, \tilde{y}$} (spline);
        \draw [arrow] (spline) -- node[midway, above] {\footnotesize State $x(t)$} (tokens);
        
        % Drop down
        \draw [arrow] (tokens) -- (norm);
        
        % Bottom Row (Right to Left flow)
        \draw [arrow] (norm) -- node[midway, above] {\footnotesize Normalized $\hat{\mathbf{t}}_k$} (transformer);
        \draw [arrow] (transformer) -- (lora);
        \draw [arrow] (lora) -- (output);
        
        % --- Physics/Noise Inputs to Spline ---
        \node [above=0.8cm of spline, align=center, font=\footnotesize] (noise) {System Dynamics \\ Measurement Models};
        \draw [arrow, dashed] (noise) -- (spline);

        % --- Model Explcit --- %
        \draw [arrow, dashed] (input) -- node[midway, right] {\footnotesize \modelname} (output);

    \end{tikzpicture}
    }
    \caption{\textbf{Continuous Tokenization Pipeline.} The raw discrete data is transformed and then enriched using optimization-based stochastic spline generation \cite{Kearney2024}. The coefficients representing Position, Velocity, Acceleration, and Jerk are extracted, anchored to the window start, and fed into a Causal Transformer.}
    \label{fig:pipeline}
\end{figure}
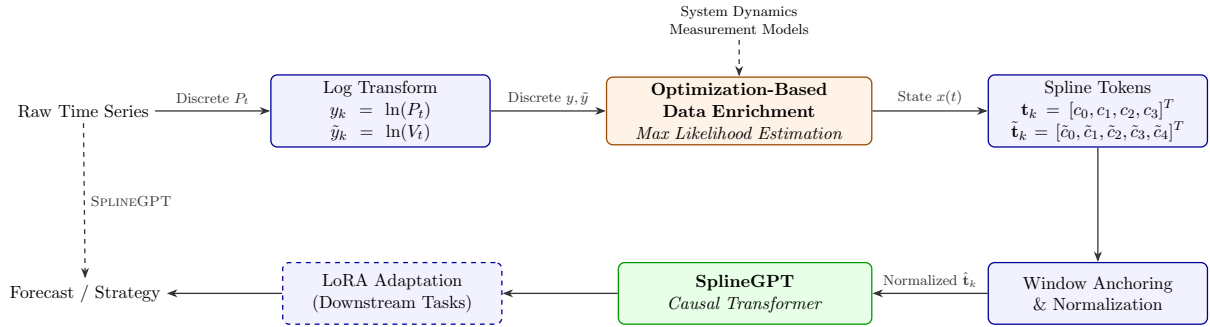

Our proposed pipeline transforms raw financial data into a rich latent representation suitable for high-dimensional embedding. The process consists of three distinct stages: (1) Signal Pre-processing, (2) Optimization-Based Spline Enrichment, and (3) Stationarity \& Normalization. We represent the proposed ML system architecture in Figure \ref{fig:pipeline}.

\subsection{Pre-tokenization Transforms}
Financial time series, specifically asset prices $P(t)$, are defined on the non-negative real line $\R^+$. Direct analysis of raw prices often introduces scaling instabilities. To map the state space to a domain better suited to standard stochastic processes, we apply a logarithmic transformation $y_k = \ln(P(t_k))$. This aligns the data with symmetric stochastic processes (e.g., Gaussian process) used in the enrichment phase.

The logarithmic transform is also applied to the daily trading volume data $V(t)$ yielding $\tilde{y}_k = \ln(V(t_k))$. As before this aligns the data with symmetric processes and promotes numerical stability by compressing dynamic range; mega-caps trade with daily volumes in the hundreds of millions whereas small-caps trade with daily volumes in the tens of thousands.

\subsection{Optimization-Based Spline Enrichment}
The core innovation relies on our previous work \cite{Kearney2024}. We view the transformed sequence $\{y_k\}$ as noisy measurements of a latent continuous state $x(t)$ governed by a stochastic differential equation (SDE).
We form splines which maximize likelihood over the interval $[t_0, t_N]$:
\begin{align}
    \label{eqn:1}
    \underset{v,w,x}{\min} & \hspace{0.25cm}\int_{t_0}^{t_N} \frac{1}{2} v^2 dt + \frac{\alpha^2}{2} \sum_k w_k^2\\
    \notag
    \mathrm{s.t.} & \hspace{0.25cm} \ddot{x}(t) = v(t)\\
    \notag
    & \hspace{0.25cm} y_k = x(t_k) + w_k, \hspace{0.25cm} k=0,1,...,N
\end{align}
Using the calculus of variations, the optimal solution $x^*(t)$ is a continuous spline defined by the ratio of process noise to measurement noise ($\alpha^2 = \frac{\sigma_p^2}{\sigma_m^2}$). The parameter $\alpha^2$ acts as a filtering parameter, separating the underlying market trend from high-frequency microstructure noise (e.g., bid-ask bounce), ensuring the tokens represent signal rather than jitter. 
In prototype models described in the present work we use a value of $\alpha = 5$ for proof of concept, but future detailed statistical studies can be used to optimize this stochastic filter parameter for improved performance.

Problem (\ref{eqn:1}) provides a rigorous kinematic basis for the tokenization of asset prices. For trading volume, we adopt a different philosophy. Since volume is an aggregate measure (shares traded \emph{over} an interval) rather than a continuous state variable, we employ a heuristic volume smoother designed to stabilize the input for the Transformer. We formulate the volume extraction as:
\begin{align}
    \label{eqn:2}
    \underset{\tilde{v},\tilde{w},\tilde{x}}{\min} & \hspace{0.25cm}\int_{t_0}^{t_N} \frac{1}{2} \tilde{v}^2 dt + \frac{\alpha^2}{2} \sum_k \tilde{w}_k^2\\
    \notag
    \mathrm{s.t.} & \hspace{0.25cm} \ddot{\tilde{x}}(t) = \tilde{v}(t)\\
    \notag
    \mathrm{s.t.} & \hspace{0.25cm} \tilde{y}_k = \int_{t_{k-1}}^{t_k} \tilde{x}(t) dt + \tilde{w}_k, \hspace{0.25cm} k= 1,...,N.
\end{align}
We explicitly distinguish the interpretation of (\ref{eqn:2}) from (\ref{eqn:1}). While (\ref{eqn:1}) represents a physical equation of motion, (\ref{eqn:2}) acts as a phenomenological filter \cite{Ramsay2005}. The measurement equation uses log-transformed aggregate volume ($\tilde{y}_k = \ln(V_k)$). Mathematically, treating the integral of the state $\tilde{x}(t)$ as consistent with the log of the sum is a known approximation. However, in the context of deep learning inputs, this heuristic serves a critical engineering purpose: it compresses the dynamic range of volume spikes (which span orders of magnitude) while enforcing temporal continuity via the spline constraints. We find this approximation empirically robust for stabilizing the tokenization of heavy-tailed volume data, preventing outlier gradients from destabilizing the Transformer backbone.

Solving (\ref{eqn:1}) generates a piecewise optimal (maximum likelihood) estimate of the log-price given by
\begin{equation}
    \label{eqn:spline-price}
    x(t) = c_0(t_k) + c_1(t_k)(t - t_k) + c_2(t_k) \frac{(t-t_k)^2}{2} + c_3(t_k)\frac{(t-t_k)^3}{6},
\end{equation}
where $t \in [t_{k},t_{k+1}]$ and the constants corresponding to this interval are provided by the optimal solution. The coefficients $c_j(t_k)$ represents the $j$-th kinematic term (e.g. position, velocity, etc) on the interval starting at $t_k$.
Similarly, solving (\ref{eqn:2}) generates a piecewise optimal estimate of the \emph{Volume Intensity Function} $\tilde{x}(t)$ defined by
\begin{equation}
    \label{eqn:spline-volume}
    \tilde{x}(t) = \tilde{c}_0(t_k) + \tilde{c}_1(t_k)(t - t_k) + \tilde{c}_2(t_k) \frac{(t-t_k)^2}{2} + \tilde{c}_3(t_k)\frac{(t-t_k)^3}{6} + \tilde{c}_4(t_k) \frac{(t - t_k)^4}{24},
\end{equation}
for $t \in [t_{k},t_{k+1}]$. Here, the coefficients $\tilde{c}$ do not represent strict physical derivatives, but rather shape parameters of the local log-volume rate profile that allow the model to detect demand surges (velocity $\tilde{c}_1$) and exhaustion patterns (acceleration $\tilde{c}_2$).

\subsection{Continuous-Time Tokens}
The solution to (\ref{eqn:1}) is a cubic spline, and therefore on each interval the spline is fully defined by $4$ coefficients. Similarly, the solution to (\ref{eqn:2}) is a quartic spline and therefore on each interval is fully defined by $5$ coefficients. From the optimal spline $x^*(t)$, we extract the coefficients defining the signal segment for each interval for use as data tokens. Unlike discrete points, our tokens $\mathbf{t}_k$ and $\tilde{\mathbf{t}}_k$ are vectors representing the kinematic state of the market on the interval:
\begin{equation}
    \label{eqn:3}
    \mathbf{t}_k = [c_0, c_1, c_2, c_3]^T \in \R^4,
\end{equation}
and
\begin{equation}
    \label{eqn:4}
    \tilde{\mathbf{t}}_k = [\tilde{c}_0, \tilde{c}_1, \tilde{c}_2, \tilde{c}_3,\tilde{c}_4]^T \in \R^5.
\end{equation}

In (\ref{eqn:3}) $c_0$ is ``position'' (log-price), $c_1$ is ``velocity'' (returns), $c_2$ is ``acceleration'' (convexity), and $c_3$ is ``jerk'' (trend instability).
This interpretation is modified in (\ref{eqn:4}) where trading volume permits interpretation as an ``energy'' equivalent; the more shares traded, the more energetic the underlying asset.
Under this analogy $\tilde{c}_0$ is ``power'' (log-volume rate), $\tilde{c}_1$ is the first time-derivative of ``power'', and so on.
This results in an architecture with a net $9$-dimensional token on each time interval when $\mathbf{t}_k$ and $\tilde{\mathbf{t}}_k$ are concatenated for processing. The concatenated token vector contains both price and volume information, enabling the Transformer to attend to coupled dynamics between the two quantities.

\subsection{Stationarity \& Normalization}
A critical challenge in training foundation models on financial data is stationarity. Raw price levels ($c_0$) and volume intensities ($\tilde{c}_0$) are non-stationary, causing ``mean reversion bias'' where the model incorrectly predicts a return to the dataset mean rather than following the local trend.

To resolve this, we implement \textbf{Dual Window Anchoring}. For every context window $W$, we subtract the initial price position $c_0(t_0)$ and initial volume intensity $\tilde{c}_0(t_0)$ from all tokens in the sequence:
\begin{align}
    c'_0(t_k) &= c_0(t_k) - c_0(t_0) \\
    \tilde{c}'_0(t_k) &= \tilde{c}_0(t_k) - \tilde{c}_0(t_0)
\end{align}
This forces every training sequence to begin at a relative origin of $(0.0, 0.0)$, allowing the model to better learn relative dynamics for both price and volume.

Furthermore, the higher-order derivatives exhibit vast scale disparities (e.g., price velocity $c_1 \approx 10^{-2}$ vs. volume jounce $\tilde{c}_4 \approx 5.7$). We compute global statistics ($\sigma \in \R^7$) for the non-anchored coefficients over the training corpus and apply Z-score normalization \cite{LeCun2012} to the derivatives ($c_1, c_2, c_3$) and volume shape parameters ($\tilde{c}_1, \tilde{c}_2, \tilde{c}_3, \tilde{c}_4$).

\section{Model Architecture and Training}

\subsection{\modelname{} Architecture Specification}
We employ a lightweight, decoder-only Causal Transformer designed for high-throughput inference and rapid adaptation. The architecture is configured as a ``Short \& Wide'' model, prioritizing high-bandwidth feature extraction (width) over deep logical reasoning (depth) to capture the complex correlations between the 9 coupled physical state variables. The specific hyperparameters are detailed in Table \ref{tab:arch}.

\begin{table}[htbp]
    \centering
    \begin{tabular}{lc}
        \toprule
        \textbf{Hyperparameter} & \textbf{Value} \\
        \midrule
        Layers ($L$) & 4 \\
        Attention Heads ($H$) & 8 \\
        Embedding Dimension ($d_{model}$) & 512 \\
        Feed-Forward Dimension ($d_{ff}$) & 2048 \\
        Context Window ($T$) & 64 \\
        Input Channels & 9 (4 Price + 5 Volume) \\
        Dropout & 0.1 \\
        Activation & GELU \\
        Positional Encoding & RoPE \\
        \bottomrule
    \end{tabular}
    \caption{\modelname{} Hyperparameters}
    \label{tab:arch}
\end{table}

The total trainable parameter count is approximately 12.6 million. While larger than typical ``Nano'' models, this capacity is beneficial for resolving the 9-dimensional kinematic state space. Despite the larger width, the shallow depth (4 layers) supports a model that is small enough for fast CPU inference / low latency deployments.
Prototype models were trained using a personal desktop with 64GB of RAM, an NVIDIA GeForce RTX 2080 SUPER GPU with 8GB of onboard memory, and an Intel i9-9900 CPU.

\subsection{Training and Testing Data Strategy}
The model was pre-trained on a large corpus of US Equities data. Following dataset consolidation and quality assurance filtering -- deficient ticker removal based on low liquidity, missing data, etc -- the final training corpus comprises 22.5 million continuous spline windows. With a context length of 64 intervals per window, this equates to approximately 1.44 Billion continuous-time tokens.

To ensure rigorous scientific validation and prevent look-ahead bias (data leakage), we enforce a strict temporal cutoff:
\begin{itemize}
    \item \textbf{Training Set (Foundation \& Adapter):} Data from Jan 1, 2000 to Dec 31, 2022.
    \item \textbf{Test Set (Evaluation Only):} Data from Jan 1, 2023 to Present.
\end{itemize}
The Foundation Model is entirely blind to the test period, ensuring that any performance on recent market regimes is a result of generalized kinematic learning rather than memorization.

\subsection{Training Stability}
Training on high-variance derivative tokens (e.g. $c_3$, $\tilde{c}_4$) presents stability challenges. To ensure convergence, we employ a \textbf{Cosine Annealing Learning Rate Scheduler}, decaying the learning rate from $6e^{-4}$ to $0$ over 5 epochs. Additionally, we implement \textbf{Gradient Clipping} (norm $\le 1.0$) to prevent exploding gradients caused by outliers in the volatility surface.

\subsection{Fine-Tuning via Low-Rank Adaptation (LoRA)}
Following the general pre-training phase, we employ Low-Rank Adaptation (LoRA) to specialize the model. LoRA freezes the pre-trained weights $W$ and introduces trainable rank decomposition matrices $A$ and $B$, such that $W' = W + BA$. We implement a strategic injection of LoRA adapters targeting the input and output processing stages of our 4-layer architecture:
\begin{enumerate}
    \item \textbf{Perception Adaptation (Layers 1 \& 2):} Fine-tunes low-level feature extraction (e.g., detecting volume-price divergence).
    \item \textbf{Reasoning Adaptation (Layer 4):} Fine-tunes the final decision boundary for strategy formulation.
\end{enumerate}
The fine-tuning head is trained for a classification objective: \texttt{Buy}, \texttt{Sell}, or \texttt{Hold}.

% -----------------------------------------
% TARGET DEFINITION
% -----------------------------------------
\subsection{Training Objectives \& Labeling}
\label{sec:labeling}
To align the model's kinematic reasoning with actionable trading signals, we employ a ternary classification objective (Buy, Sell, Hold) based on a ``Momentum Logic'' threshold. We adopt the notation $g_j$ for the discrete categorical label to distinguish it from the measurement observations $y_k$ used in the enrichment phase.

For a given context window ending at time $t_j$, the label $g_j$ is determined by the state of the target token $c_0(t_{j+1})$ relative to the last input token $c_0(t_j)$:
\begin{equation}
    g_j = 
    \begin{cases} 
      0 (\text{Buy}) & \text{if } r_j > \tau \\
      1 (\text{Sell}) & \text{if } r_j < -\tau \\
      2 (\text{Hold}) & \text{otherwise}
    \end{cases}
\end{equation}
where $r_j = c_0(t_{j+1}) - c_0(t_j)$ is the log-return over the next interval. In this work, we set the volatility threshold $\tau = 0.01$ (1\%). This filters out minor noise, forcing the model to only predict ``actionable'' moves when the kinematic signal suggests a significant price excursion.

To induce the ``Active Risk Management'' behavior, we train using an asymmetric Weighted Cross-Entropy Loss:
\begin{equation}
    \mathcal{L} = -\sum_{i \in \{0,1,2\}} w_i \mathbb{I}(g=i) \log(\hat{g}_i)
\end{equation}
Here, $\hat{g}_i$ represents the predicted probability assigned by the model to class $i$, and $\mathbb{I}(g=i)$ is the indicator function for the true label $g$. We assign weights $\mathbf{w} = [2.0, 10.0, 1.0]$. The high penalty on the Sell class ($w_{sell}=10.0$) conditions the model to be hyper-sensitive to downside kinematic signatures (e.g., negative jerk), prioritizing capital preservation over accuracy.

% -----------------------------------------------------------
% BASELINE DEFINITIONS
% -----------------------------------------------------------
\subsection{Tokenization Ablations and External Baseline}
To isolate the effect of continuous kinematic tokenization, we perform two complementary comparisons.
\textbf{(i) Controlled internal ablations:} we hold the Transformer backbone, parameter count, training objective, and adaptation method fixed, varying only the input representation (Spline vs. discrete Raw vs. discrete finite differences).
\textbf{(ii) External architecture baseline:} we additionally compare against PatchTST \cite{nie2023timeseriesworth64} as a representative patching-based time-series foundation model. This second comparison matches parameter scale but is \emph{not} strictly ceteris paribus, and is therefore interpreted as a behavioral robustness reference rather than a controlled ablation.

\begin{table}[htbp]
    \centering
    \resizebox{\textwidth}{!}{%
    \begin{tabular}{|lccccc|}
        \toprule
        \textbf{Model} & \textbf{Input Representation} & \textbf{Physics?} & \textbf{Backbone} & \textbf{Params} & \textbf{Adaptation} \\
        \midrule
        \textbf{\modelname{}} & \textbf{Continuous Spline Coeffs} & \textbf{Explicit (Exact)} & \textbf{Transf. (4L, 8H)} & \textbf{12.6M} & \textbf{LoRA} \\
        RawGPT-FD & Discrete Finite Differences & Explicit (Noisy) & Transf. (4L, 8H) & 12.6M & LoRA \\
        RawGPT & Raw Discrete Values & None (Ablated) & Transf. (4L, 8H) & 12.6M & LoRA \\
        \midrule
        PatchTST \cite{nie2023timeseriesworth64} & Discrete Patches ($P=16, S=8$) & Implicit (Latent) & Transf. (4L, 4H) & $\approx$12M & Linear Probe \\
        \bottomrule
    \end{tabular}%
    }
    \caption{\textbf{Model Specifications.} \textbf{Controlled ablations} (SplineGPT, RawGPT-FD, RawGPT) share the same Transformer backbone and training objective; differences are isolated to input representation. \textbf{PatchTST} is included as an \emph{external} patching-based reference model with comparable parameter scale, but differs architecturally and in adaptation method.}
\end{table}

\subsubsection{Internal Ablations: The Value of Smoothness}
We define two internal variants that utilize the exact same transformer backbone as \modelname{} but lack the spline enrichment step.

\paragraph{RawGPT-FD (The Discrete Physics Baseline):}
This model tests whether the \emph{smoothness and accuracy} of the spline is superior to standard discrete differentiators. It is provided with the full kinematic state vector (Position, Velocity, Acceleration), but these features are derived via explicit finite differences:
\begin{equation}
    c_{n,k} \approx \Delta^n y_k = \sum_{i=0}^n (-1)^{n-i} \binom{n}{i} y_{k-(n-i)}.
\end{equation}
While this theoretically provides the same information as the spline ($c_2, c_3$), discrete differentiation amplifies measurement noise ($\text{Var}(\Delta y) \approx 2\text{Var}(y)$). If \modelname{} outperforms RawGPT-FD, it confirms the value of \emph{optimization-based denoising} over arithmetic approximation.

\paragraph{RawGPT (The Physics-Ablated Baseline):}
This model represents an information-poor condition. It tests whether high-order kinematics are necessary at all. We map discrete inputs directly to the channel dimension, but we mask the higher-order derivatives ($c_2, c_3$) and volume shape parameters ($\tilde{c}_2-\tilde{c}_4$) with zeros. This forces the model to rely solely on raw directional moves and scalar volatility, effectively removing the ``physics'' from the input.

\subsubsection{External Benchmark: PatchTST}
To validate that our results are not artifacts of our specific backbone, we benchmark against \textbf{PatchTST} \cite{nie2023timeseriesworth64}, a state-of-the-art foundation model for multivariate time series.
\begin{itemize}
    \item \textbf{Matched Capacity:} We configure the PatchTST backbone (Channel-Independent) with $d_{model}=256$ and 4 heads. This results in a total parameter count ($\approx 12M$) comparable to \modelname{}, preventing the benchmark from being under-parameterized.
    \item \textbf{Implicit Smoothing:} PatchTST relies on patch-based tokenization ($P=16, S=8$) to capture local semantics. This serves as a test of \emph{Implicit} vs. \emph{Explicit} smoothing: does the model learn better when it aggregates noisy points itself (Patching), or when we mathematically pre-process the points into smooth curves (Splines)?
\end{itemize}
Accordingly, PatchTST results are interpreted as an external reference point rather than a strictly controlled ablation.

% -----------------------------------------
% EXPERIMENTAL SETUP
% -----------------------------------------
\section{Experimental Validation}

We conducted a backtesting campaign on a test set comprised of 6 major US equities (INTC, NVDA, JPM, XOM, PFE, TSLA) from Jan 1, 2023, to Present.

\subsection{Experimental Setup \& Protocol}
A critical requirement for valid time-series backtesting is the strict prevention of look-ahead bias. To ensure this, we implement a \textbf{Rolling Spline Tokenization} protocol. We do \textit{not} fit a single global spline to the entire dataset. Instead, for every inference step $t$:
\begin{enumerate}
    \item \textbf{Window Extraction:} We retrieve the raw discrete price/volume history strictly for the window $[t - T, t]$.
    \item \textbf{Local Enrichment:} We solve the optimization problems (\ref{eqn:1}) and (\ref{eqn:2}) on this isolated window to generate the continuous function $x^*(t)$.
    \item \textbf{Tokenization:} We extract the coefficients $\mathbf{t}_{t-T:t}$ from this local spline.
    \item \textbf{Inference:} The Transformer processes these tokens to predict the state at $t+1$.
\end{enumerate}
The parameter $T$ denotes the context window length. This ensures that the ``kinematic state'' at time $t$ is derived solely from information available up to time $t$.

\subsubsection{Backtest Constraints \& Limitations}
The simulation assumes an initial capital of \$10,000. We enforce a tax penalty by applying a 32\% short-term capital gains tax on realized profits at the end of every 252-day trading year. 
\begin{itemize}
    \item \textbf{Execution:} Trades are simulated at the daily close price at time $t$, coincident with the inference step. We assume a ``Market-On-Close'' (MOC) execution regime where the model inference is completed using near-close data, and orders are filled at the closing auction. Given the high liquidity of the selected large-cap universe and the experimental focus on representation learning rather than high-turnover arbitrage, we treat execution latency as negligible.
    \item \textbf{Limitations:} This study isolates the predictive power of the representation; therefore, it does not currently model variable transaction costs (slippage, commissions) or market impact. While the notably profitable performance of the model provides a significant buffer against these costs, future work will incorporate full order book dynamics.
    \item \textbf{Baselines:} The primary financial benchmark is the Buy \& Hold strategy, representing the standard opportunity cost for an investor. 
\end{itemize}

\subsubsection{Trading Policy \& State Machine}
To translate the ternary model outputs ($g_t \in \{\text{Buy, Sell, Hold}\}$) into portfolio states, we implement a strict finite state machine (FSM). The portfolio state $S_t$ at time $t$ is binary: $S_t \in \{0 \text{ (Cash)}, 1 \text{ (Long)}\}$. Short selling and leverage are explicitly disabled to isolate the signal quality from leverage effects.

The state transitions are governed by the following logic:
\begin{itemize}
    \item \textbf{Entry (Cash $\to$ Long):} If $S_t = 0$ and the model predicts $g_t = \text{Buy}$, a buy order is executed at the closing price $P_t$. The entire available capital (minus transaction costs) is converted to shares.
    \item \textbf{Exit (Long $\to$ Cash):} If $S_t = 1$ and the model predicts $g_t = \text{Sell}$, a sell order is executed at $P_t$. All held shares are liquidated to cash.
    \item \textbf{Maintenance:} If $g_t = \text{Hold}$, or if the model signals an invalid transition (e.g., signaling ``Buy'' when already Long), the state remains unchanged ($S_{t+1} = S_t$).
\end{itemize}
\paragraph{Definition (Liquidation Equilibrium).}
We distinguish \emph{action-space} outputs from \emph{portfolio-state} outcomes.
A model may collapse to a degenerate action policy (e.g., predicting a single label).
Under our FSM, any degenerate policy that fails to generate valid entries (Buy while in Cash) yields an \emph{absorbing cash state} $S_t=0$ with zero exposure.
We term this portfolio-level absorbing behavior the \textbf{Liquidation Equilibrium}.
In our experiments, the observed Liquidation Equilibrium for discrete baselines is realized specifically through a monotonic \textbf{Always-Sell} prediction ($g_t=\text{Sell}$), which immediately (or persistently) drives and maintains $S_t=0$.
This definition clarifies that the ``Liquidation Equilibrium'' observed in the baselines (Section \ref{sec:noise_wall}) corresponds to the absorbing state $S_t=0$, reached when a model monotonically predicts $g_t=\text{Sell}$ regardless of input.

\subsubsection{Evaluation Metrics}
To rigorously quantify the risk-adjusted performance of the learned representation, we report the following metrics alongside total return:

\begin{itemize}
    \item \textbf{Sharpe Ratio:} Measures excess return per unit of total volatility. Defined as 
    \[
        S = \frac{E[R_p - R_f]}{\eta},
    \] 
    where $R_p$ is the annualized portfolio return, $R_f$ is the risk-free rate (assumed $4\%$ for the 2023--2025 period), and $\eta$ is the annualized standard deviation of daily returns.
    \item \textbf{Sortino Ratio:} A modification of the Sharpe ratio that penalizes only downside volatility. Defined as 
    \[
        S_{ort} = \frac{E[R_p - R_f]}{\sigma_d},
    \] 
    where $\sigma_d$ is the standard deviation of negative asset returns (downside deviation). This metric is particularly relevant for our ``Active Risk Management'' objective, which seeks to minimize capital destruction rather than variance per se.
    \item \textbf{Maximum Drawdown (MaxDD):} The maximum observed loss from a peak to a trough of the portfolio equity curve over the test period: 
    \[
        \mathrm{MaxDD} = \min_{t} \left( \frac{C_t - \max_{s \le t} C_s}{\max_{s \le t} C_s} \right),
    \]
    where $C_t$ is the value of the portfolio at time $t$.
    This quantifies the worst-case scenario for an investor holding the strategy.
\end{itemize}

\subsection{Results}
\label{sec:results}

We present our findings in three stages: (1) Behavioral Diagnostics, establishing that the model is actually learning a policy rather than collapsing; (2) Financial Performance, quantifying the value of that policy; and (3) Robustness, verifying stability across thresholds and costs.

\subsubsection{Behavioral Diagnostics: Escaping the Liquidation Equilibrium}
Before evaluating returns, we strictly evaluate the \emph{learnability} of the task. Given the asymmetric loss ($w_{sell}=10.0$), a model with low signal confidence will rationally converge to a ``Liquidation Equilibrium'' (always Sell/Cash) to minimize risk.

\begin{figure}[htbp]
    \centering
    \includegraphics[width=\textwidth]{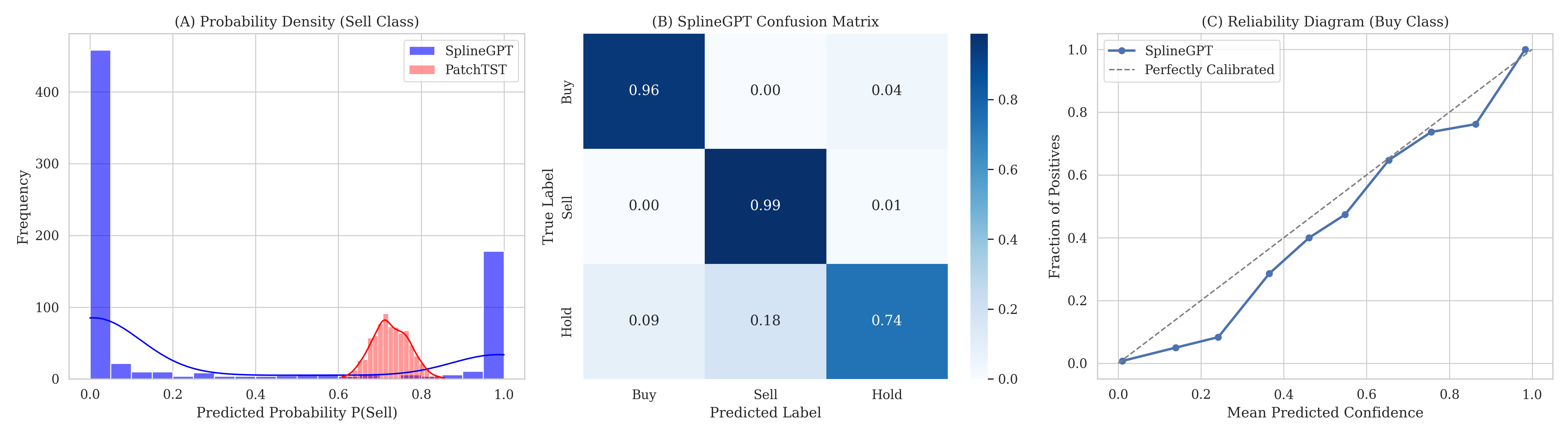}
    \caption{\textbf{Behavioral Divergence and Calibration (NVDA).} \textbf{(A)} Action distribution shows the discrete baseline (PatchTST, Red) collapsing to a narrow density peak (Liquidation Equilibrium), whereas \modelname{} (Blue) maintains a broad, active policy distribution. \textbf{(B)} Confusion matrix confirms \modelname{} achieves meaningful separation between Hold (2) and Action (0/1) states, rather than defaulting to a single class. \textbf{(C)} Calibration plot indicates the model's probability estimates for ``Buy'' signals track the empirical win rate, staying close to the diagonal (perfect calibration).}
    \label{fig:diagnostics}
\end{figure}

Figure \ref{fig:diagnostics} visualizes this divergence for the NVDA trading study. While the discrete baseline (PatchTST) collapses to a monotonic distribution (Action Rate $\to$ 100\% Sell), \modelname{} maintains a balanced ternary distribution. This confirms that the continuous spline representation provides sufficient signal-to-noise ratio to discern actionable information and justify entry risk.

\subsubsection{Financial Performance}
Having established behavioral validity, we report the financial outcomes in Tables \ref{tab:absolute_perf} and \ref{tab:risk_adjusted}, with cumulative equity curves visualized in Figure \ref{fig:sector_sweep}.
\modelname{} reliably outperforms the market benchmark on downside risk (Max Drawdown) while capturing significant upside in momentum regimes (NVDA, TSLA). Notably, in the PFE crash scenario, the model's ability to recognize negative kinematic signatures (Jerk) allowed it to exit early, resulting in a flat equity curve while the asset declined $\approx 40\%$.

\begin{table}[htbp]
    \centering
    \resizebox{\textwidth}{!}{%
    \begin{tabular}{|l|ccccc|ccccc|}
        \toprule
        & \multicolumn{5}{c|}{\textbf{Total Return (Cumulative)}} & \multicolumn{5}{c}{\textbf{Max Drawdown (Risk)}} \\
        \midrule
        \textbf{Ticker} & \textbf{Spline} & \textbf{Patch} & \textbf{FD} & \textbf{Raw} & \textbf{Mkt} & \textbf{Spline} & \textbf{Patch} & \textbf{FD} & \textbf{Raw} & \textbf{Mkt} \\
        \midrule
        \textbf{INTC} & \textbf{+240.4\%} & 0.0\% & 0.0\% & 0.0\% & +71.6\% & \textbf{-34.7\%} & 0.0\% & 0.0\% & 0.0\% & -48.2\% \\
        \textbf{NVDA} & +579.2\% & 0.0\% & 0.0\% & 0.0\% & +1185.7\% & \textbf{-32.1\%} & 0.0\% & 0.0\% & 0.0\% & -56.4\% \\
        \textbf{JPM}  & +82.0\% & 0.0\% & 0.0\% & 0.0\% & +159.9\% & \textbf{-12.5\%} & 0.0\% & 0.0\% & 0.0\% & -22.1\% \\
        \textbf{XOM}  & +48.4\% & 0.0\% & 0.0\% & 0.0\% & +28.5\% & \textbf{-13.5\%} & 0.0\% & 0.0\% & 0.0\% & -26.3\% \\
        \textbf{PFE}  & \textbf{+3.6\%} & 0.0\% & 0.0\% & 0.0\% & -41.2\% & \textbf{-18.8\%} & 0.0\% & 0.0\% & 0.0\% & -44.5\% \\
        \textbf{TSLA} & \textbf{+699.8\%} & 0.0\% & 0.0\% & 0.0\% & +306.2\% & \textbf{-33.4\%} & 0.0\% & 0.0\% & 0.0\% & -63.5\% \\
        \bottomrule
    \end{tabular}
    }
    \caption{\textbf{Absolute Performance Metrics.} All non-spline baselines collapsed to a \textbf{Liquidation Equilibrium} (0.0\% return/risk). \modelname{} generates active policies with superior capital preservation (MaxDD).}
    \label{tab:absolute_perf}
\end{table}

\begin{table}[htbp]
    \centering
    \resizebox{\textwidth}{!}{%
    \begin{tabular}{|l|ccccc|ccccc|}
        \toprule
        & \multicolumn{5}{c|}{\textbf{Sharpe Ratio}} & \multicolumn{5}{c}{\textbf{Sortino Ratio}} \\
        \midrule
        \textbf{Ticker} & \textbf{Spline} & \textbf{Patch} & \textbf{FD} & \textbf{Raw} & \textbf{Mkt} & \textbf{Spline} & \textbf{Patch} & \textbf{FD} & \textbf{Raw} & \textbf{Mkt} \\
        \midrule
        \textbf{INTC} & \textbf{1.24} & N/A & N/A & N/A & 0.85 & \textbf{1.13} & N/A & N/A & N/A & 1.12 \\
        \textbf{NVDA} & 2.31 & N/A & N/A & N/A & 2.80 & 2.16 & N/A & N/A & N/A & 3.95 \\
        \textbf{JPM}  & 0.98 & N/A & N/A & N/A & 1.60 & 0.88 & N/A & N/A & N/A & 2.25 \\
        \textbf{XOM}  & \textbf{0.61} & N/A & N/A & N/A & 0.55 & 0.61 & N/A & N/A & N/A & 0.78 \\
        \textbf{PFE}  & \textbf{-0.19} & N/A & N/A & N/A & -0.80 & \textbf{-0.18} & N/A & N/A & N/A & -1.05 \\
        \textbf{TSLA} & \textbf{2.20} & N/A & N/A & N/A & 1.15 & \textbf{2.23} & N/A & N/A & N/A & 1.65 \\
        \bottomrule
    \end{tabular}
    }
    \caption{\textbf{Risk-Adjusted Efficiency.} Risk-adjusted metrics are undefined for strategies in the Liquidation Equilibrium. Under the asymmetric objective, PatchTST, FD, and Raw converge to this equilibrium (realized via monotonic Always-Sell predictions), yielding N/A Sharpe/Sortino. \modelname{} yields non-trivial action and measurable efficiency.}
    \label{tab:risk_adjusted}
\end{table}

\subsubsection{Robustness \& Sensitivity}
\begin{figure}[htbp]
    \centering
    \includegraphics[width=\textwidth]{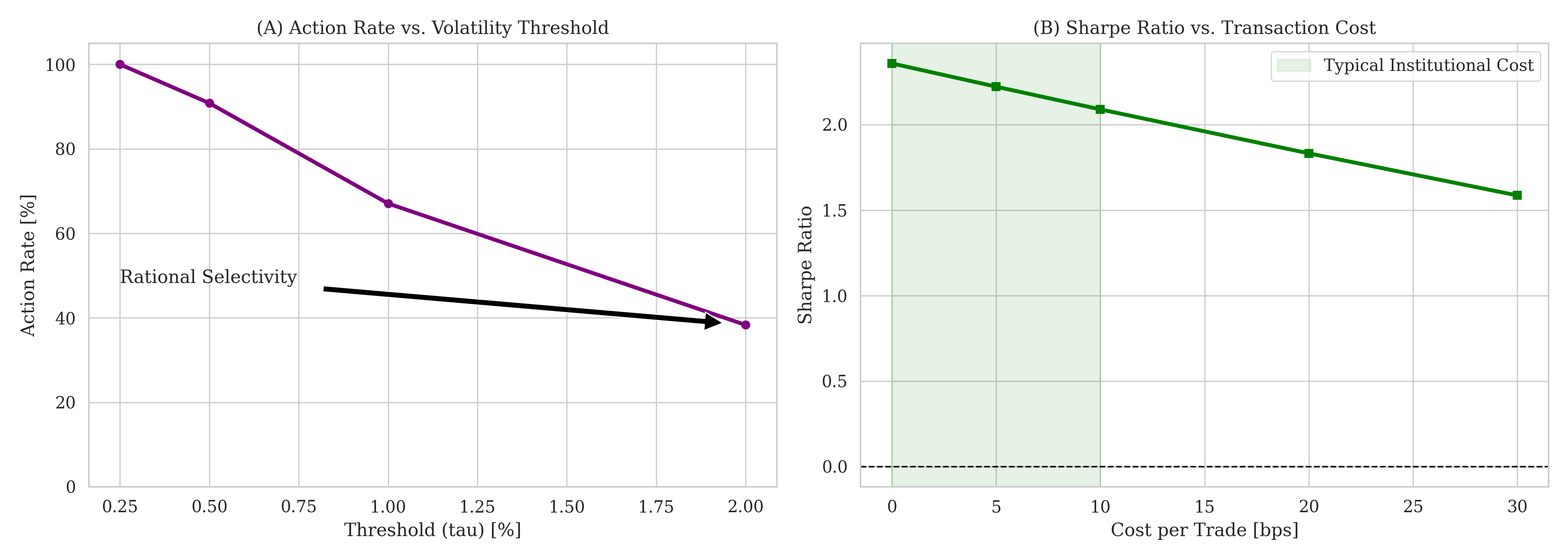}
    \caption{\textbf{Parameter Sensitivity Analysis (NVDA).} \textbf{(A)} \modelname{} exhibits \emph{Rational Selectivity}: as the volatility threshold ($\tau$) increases (making the task harder), the model monotonically reduces its action rate, avoiding false positives. \textbf{(B)} Performance remains robust (Sharpe $> 2.0$) even under high-friction regimes (up to 30bps), validating that the alpha generated by the kinematic tokens is not merely microstructure noise arbitrage.}
    \label{fig:sweeps}
\end{figure}

To further validate that these results are not artifacts of a specific threshold ($\tau=1\%$) or idealized execution, we perform comprehensive parameter sweeps using NVDA as a case study. Figure \ref{fig:sweeps} demonstrates that \modelname{} maintains positive expectancy across a wide range of transaction costs (up to 30bps) and volatility thresholds. Conversely, the discrete baselines fail to activate even when the threshold $\tau$ is lowered to $0.25\%$, indicating a fundamental representational failure rather than a tuning mismatch.

% --- PLACEHOLDER 2: SWEEPS ---

\subsubsection{Kinematic Responsiveness \& Regime Adaptation}
\label{sec:responsiveness}
\begin{figure}[p]
    \vspace*{-1.5cm}
    \centering
    % --- FIGURE 1 ---
    \includegraphics[width=0.85\textwidth]{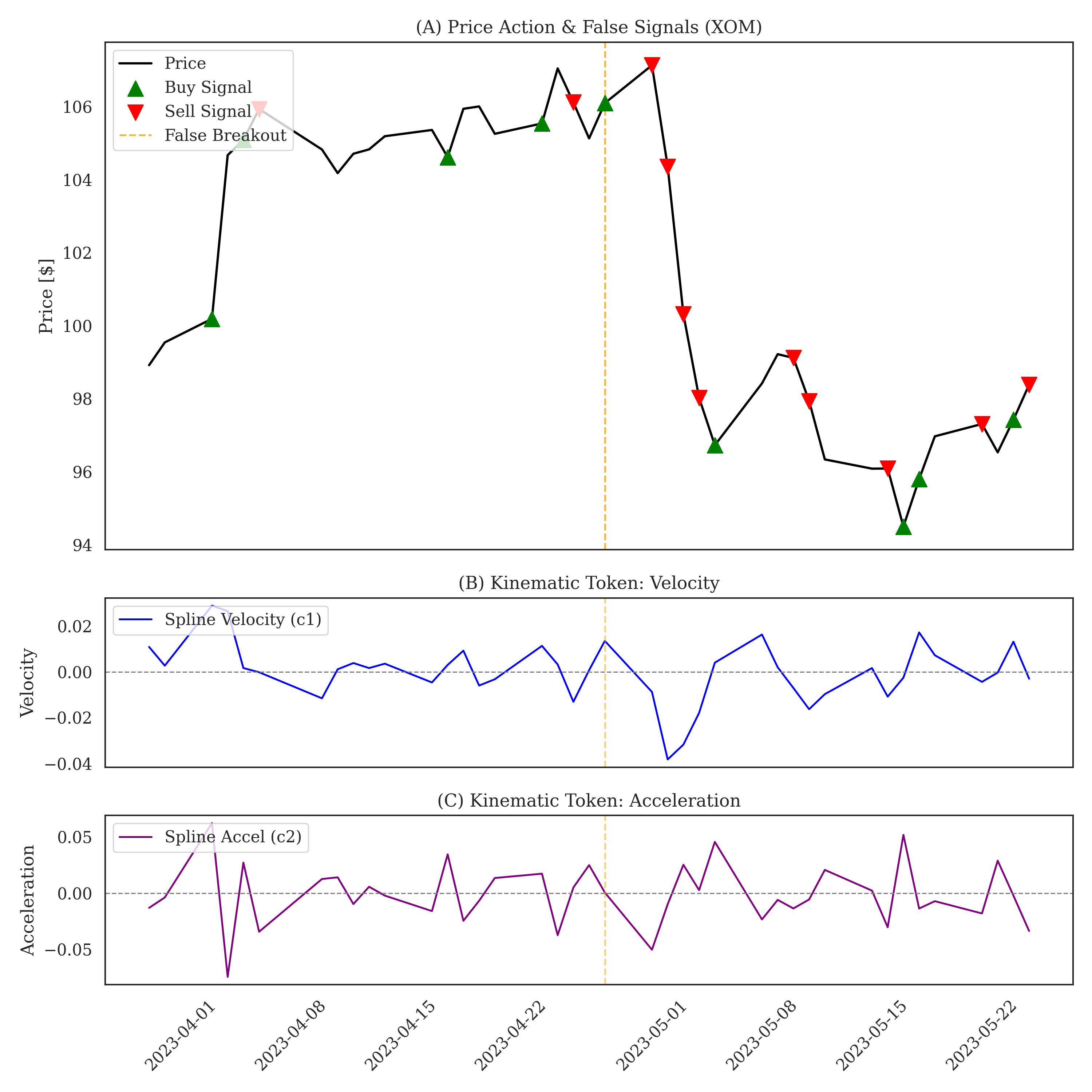}
    \caption{\textbf{Kinematic Responsiveness (XOM).} Rather than a failure, this false breakout illustrates the model's rapid adaptation. Although high velocity ($c_1$) triggered an initial entry, the immediate rollover in acceleration ($c_2$, Panel C) alerted the model to the failed breakout. The policy flipped to "Sell" (Red Triangles) within days, protecting capital from the subsequent $\approx 10\%$ decline. This rapid reaction speed explains the low Max Drawdown metrics relative to the market.}
    \label{fig:responsiveness}

    \vspace{0.8cm} % Adds vertical breathing room between the plots

    % --- FIGURE 2 ---
    \includegraphics[width=\textwidth]{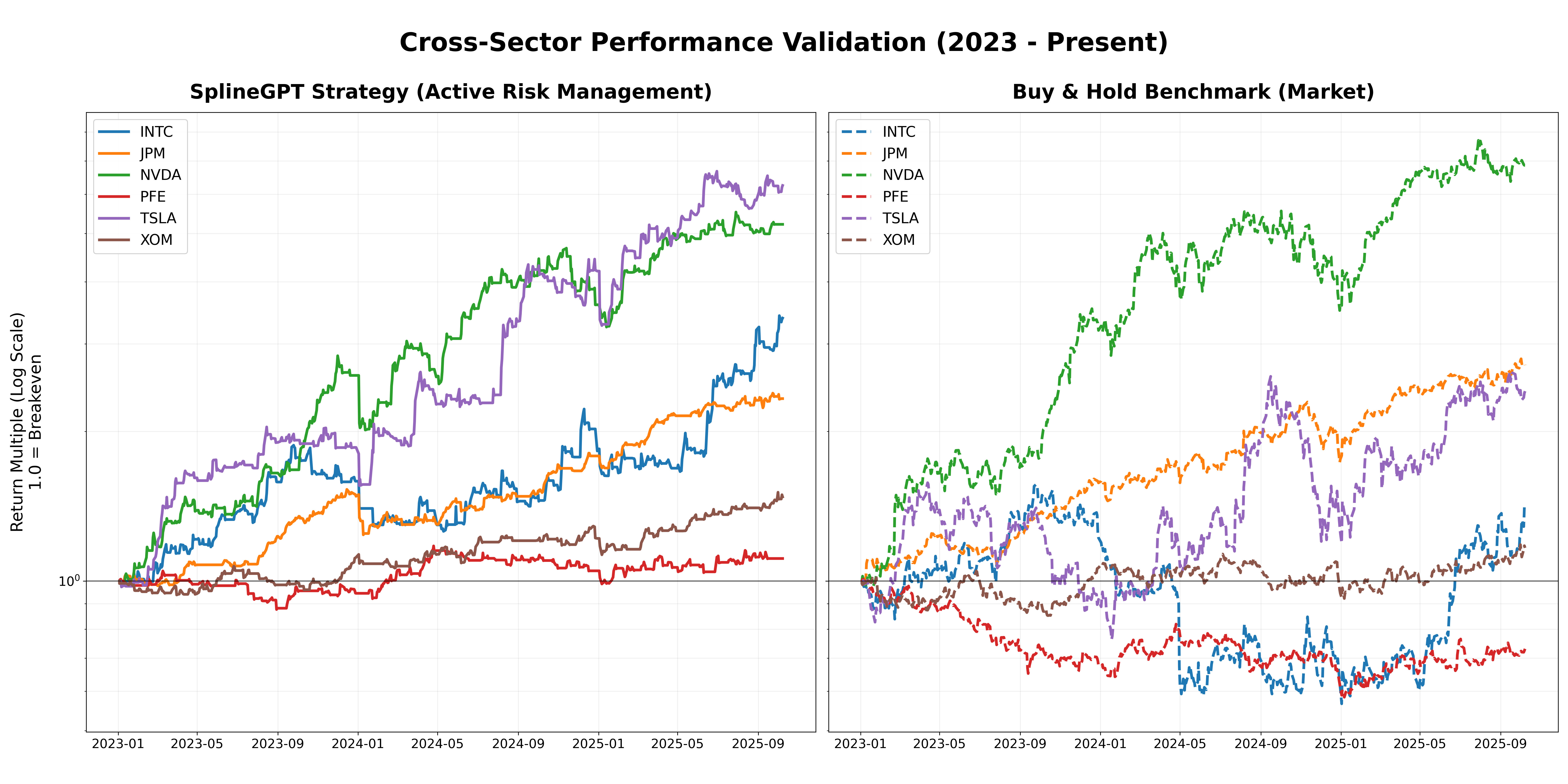}
    \caption{\textbf{Cross-Sector Performance Validation (2023 -- Present).} 
    \textbf{Left Panel:} Equity curves generated by the \modelname{} strategy using Active Risk Management weights. 
    \textbf{Right Panel:} The Buy \& Hold benchmark for the same assets. 
    Note the strategy's ability to smooth out volatility (TSLA) and strictly preserve capital during the PFE crash (red line), yielding a flat return profile while the asset declined 40\%. 
    The ablation baselines are omitted because they converged to the \emph{Liquidation Equilibrium}, realized here by a monotonic \textbf{Always-Sell} action policy; see Tables \ref{tab:absolute_perf}, \ref{tab:risk_adjusted} and \ref{tab:liquidation_equilibrium}.}
    \label{fig:sector_sweep}
\end{figure}

Finally, we examine the mechanism responsible for the superior downside protection observed in Table \ref{tab:absolute_perf}. Figure \ref{fig:responsiveness} visualizes a high-frequency "whipsaw" scenario in XOM.

Crucially, while the model triggers a momentum entry (Green Triangle) on a breakout attempt, it does not blindly hold when the move fails. As visualized in Panels B and C, the rapid decay in the Acceleration token ($c_2$) acts as a leading indicator of trend exhaustion. The model detects this kinematic breakdown and flips to a "Sell" state (Red Triangles) \emph{before} the primary price collapse occurs. This confirms that the higher-order derivative tokens allow the policy to distinguish between healthy momentum and stalling price action faster than price-lagged indicators.

% -----------------------------------------------------------
%  DISCUSSION
% -----------------------------------------------------------
\section{Discussion}
\label{sec:discussion}

The results provide evidence supporting the Physics-Informed Tokenization hypothesis in this daily-equity testbed. By benchmarking \modelname{} against a hierarchy of baselines (physics-ablated (RawGPT), discrete finite-difference approximations (RawGPT-FD), and patch-based aggregation (PatchTST)), we observe a consistent qualitative separation: under an asymmetric action objective, the baselines converge to an abstention (``Hold'') equilibrium, whereas \modelname{} learns a stable selective trading policy. This separation is central to our claim: the contribution is not only predictive accuracy, but the \emph{learnability and stability} of non-trivial behavior in a regime where abstention is rational.

\subsection{Signal Deficits and Liquidation Equilibrium}
\label{sec:noise_wall}

A critical finding is the behavioral collapse of the discrete baselines. While the absolute return metrics (Table \ref{tab:absolute_perf}) showed 0.0\% for PatchTST, the diagnostic analysis (Figure \ref{fig:diagnostics}) reveals this was due to a convergence to the Liquidation Equilibrium (100\% Sell/Cash).

This phenomenon is a rational response to the asymmetric loss function ($w_{sell}=10.0$). To trigger a ``Buy'' action, the model's posterior probability of a price rise must exceed the risk-adjusted threshold implied by the penalty. The collapse indicates that standard discrete tokenization imposes a \textbf{signal confidence barrier}: the discrete input features do not provide sufficient information gain to overcome the prior probability of loss, forcing the optimizer to default to the global risk-minimizing state (Cash).
In contrast, \modelname{} maintains a balanced active policy. This implies that the continuous kinematic tokens (particularly velocity and acceleration) successfully extract sufficient actionable signal to justify risk exposure, effectively resolving the latent structure that remains opaque to discrete representations.

\begin{table}[htbp]
    \centering
    \resizebox{0.9\textwidth}{!}{%
    \begin{tabular}{|c|ccc|ccc|}
        \toprule
        \textbf{Model} & \multicolumn{3}{c|}{\textbf{Action Distribution}} & \multicolumn{3}{c}{\textbf{Performance}} \\
        & \textbf{Buy} & \textbf{Sell} & \textbf{Hold} & \textbf{Action Rate} & \textbf{Return} & \textbf{Sharpe} \\
        \midrule
        \textbf{PatchTST} & 0 & \textbf{762} & 0 & 100.0\% & 0.0\% & N/A$^*$ \\
        \textbf{\modelname{}} & \textbf{294} & 228 & 240 & 68.5\% & \textbf{+579.2\%} & \textbf{2.31} \\
        \bottomrule
    \end{tabular}%
    }
    \caption{\textbf{The Liquidation Equilibrium of Discrete Tokens.} We diagnosed the specific action distribution of the baseline (PatchTST) versus \modelname{} at the standard volatility threshold ($\tau=1.0\%$) for NVDA. While \modelname{} maintains a balanced trading policy, the discrete baseline collapses to a monotonic ``Always Sell'' policy. This indicates that the discrete representation fails to generate sufficient signal confidence to overcome the asymmetric risk penalty ($w_{sell}=10.0$), forcing the model into the safest possible state (100\% Cash).}\footnotesize{$^*$Sharpe ratio is undefined for static cash positions (zero volatility).}
    \label{tab:liquidation_equilibrium}
\end{table}

\subsubsection{Sensitivity Sweep Analysis}

To verify that the baseline failure was not merely a result of an overly conservative threshold, we performed a sensitivity sweep, lowering the volatility threshold to $\tau=0.5\%$ to encourage model activity.
As shown in Table \ref{tab:sensitivity_sweep}, the discrete baseline failed to recover even in this aggressive regime, remaining trapped in the Liquidation Equilibrium. Conversely, \modelname{} adapted successfully, increasing its action rate to $90.8\%$ while improving its Sharpe ratio to $2.48$ and better capturing the extreme bullish nature of the underlying NVDA equity by being actively invested more often. This demonstrates that the performance gap is representational where discrete measurement tokens (versus discrete spline tokens) lack the signal-to-noise ratio required to support non-trivial decision boundaries under this standard PatchTST architecture, regardless of the threshold.

\begin{table}[htbp]
    \centering
    \resizebox{0.95\textwidth}{!}{%
    \begin{tabular}{|c|cc|cc|cc|}
        \toprule
        \textbf{Condition} & \multicolumn{2}{c|}{\textbf{PatchTST}} & \multicolumn{2}{c|}{\textbf{\modelname{}}} & \multicolumn{2}{c}{\textbf{Divergence}} \\
        ($\tau$ Threshold) & \textbf{Action Rate} & \textbf{Sharpe} & \textbf{Action Rate} & \textbf{Sharpe} & \textbf{Behavior} & \textbf{Outcome} \\ \hline
        \textbf{Very Aggressive} & \multirow{4}{*}{\vspace{-1.5cm}100.0\% (Sell)} & \multirow{4}{*}{\vspace{-1.5cm}N/A} & 100.0\% & \textbf{2.73} & Active & High-Yield \\
        ($\tau=0.25\%$) & & & & & & \\ \cline{1-1} \cline{4-7}
        \textbf{Aggressive} & & & 90.8\% & \textbf{2.48} & Active & High-Yield \\
        ($\tau=0.50\%$) & & & & & & \\ \cline{1-1} \cline{4-7}
        \textbf{Conservative} & & & 67.1\% & \textbf{2.20} & Selective & Robust \\
        ($\tau=1.00\%$) & & & & & & \\ \cline{1-1} \cline{4-7}
        \textbf{Very Conservative} & & & 38.3\% & \textbf{1.74} & Precision & Defensive \\
        ($\tau=2.00\%$) & & & & & & \\ \hline
        \bottomrule
    \end{tabular}%
    }
    \caption{\textbf{Behavioral Divergence Under Sensitivity Sweep (NVDA).} We varied the volatility threshold ($\tau$) from 0.25\% to 2.0\%. While \modelname{} scaled its activity rationally—from 100\% activity in aggressive regimes to 38\% precision in conservative ones—the discrete baseline (PatchTST) remained trapped in the Liquidation Equilibrium (100\% Sell) across all thresholds. In the most aggressive regime \modelname{} also exhibits a 100\% action rate, but this includes a mix of both buy and sell decisions. This confirms that the baseline's failure is a fundamental signal deficit, not a tuning artifact.}
    \label{tab:sensitivity_sweep}
\end{table}

\subsection{Explicit vs. Implicit Smoothing}

PatchTST is primarily optimized for continuous forecasting under an MSE objective, whereas our downstream task is a discrete, asymmetric action classification problem; thus this comparison should be interpreted as a \emph{behavioral robustness} test rather than a pure forecasting benchmark.
One plausible driver is Channel Independence (CI): PatchTST processes Price and Volume as isolated univariate series, while \modelname{} couples them into a unified 9-dimensional kinematic token.

\subsection{Objective-Induced Conservatism}

In parabolic regimes (e.g., NVDA, JPM), \modelname{} yields positive returns but trails the Buy \& Hold benchmark. This divergence is not a failure of prediction, but a structural consequence of the asymmetric risk objective (10).
Standard trend-following optimizes for total return ($R$), implicitly accepting infinite drawdown risk. In contrast, our objective functions as a constrained optimization problem: maximizing $R$ subject to strict downside penalties. Consequently, the model exhibits convexity shedding—exiting positions during intermediate mean-reversion events to reset risk exposure. This behavior characterizes the model as a ``regime-switching'' agent rather than a pure momentum factor, trading absolute ceiling for a higher risk-adjusted floor (Sharpe/Sortino).

\subsection{Sensitivity to Execution Costs \& Turnover}

To address the practical realizability of the generated signals, we conducted a sensitivity analysis across varying transaction cost regimes ranging from 0 to 20 basis points (bps) per trade. Table \ref{tab:sensitivity} summarizes the risk-adjusted returns (Sharpe Ratio) and annualized turnover for each asset.

We observe high annualized turnover (36x--89x), indicating that the model has learned a short-horizon active trading policy rather than a passive holding strategy. Despite this high-turnover behavior, the strategy demonstrates significant resilience to friction:
\begin{itemize}
    \item \textbf{High-Volatility Assets (NVDA, TSLA):} Retain strong risk-adjusted performance (Sharpe $>$ 0.49) even under the aggressive 20bps cost assumption, confirming that the kinematic signal captures price moves large enough to overcome substantial friction.
    \item \textbf{Lower-Volatility Assets (XOM, JPM):} Exhibit higher sensitivity. While JPM remains robust, XOM's edge decays near the 10bps threshold. This aligns with the expectation that spline-based derivatives provide the most value in regimes where "physical" forces (momentum, acceleration) are pronounced.
\end{itemize}

\begin{table}[htbp]
\centering
\resizebox{\textwidth}{!}{%
\begin{tabular}{|l|cccc|c|}
\toprule
\textbf{Ticker} & \textbf{Sharpe (0 bps)} & \textbf{Sharpe (5 bps)} & \textbf{Sharpe (10 bps)} & \textbf{Sharpe (20 bps)} & \textbf{Ann. Turnover} \\
\midrule
\textbf{INTC} & 1.31 & 1.19 & 1.06 & 0.83 & 77.7x \\
\textbf{NVDA} & 2.39 & 2.26 & 2.12 & 1.87 & 70.3x \\
\textbf{JPM}  & 1.15 & 1.07 & 0.99 & 0.82 & 36.0x \\
\textbf{XOM}  & 0.73 & 0.62 & 0.50 & 0.27 & 47.6x \\
\textbf{PFE}  & -0.31 & -0.43 & -0.56 & -0.83 & 44.6x \\
\textbf{TSLA} & 2.05 & 1.90 & 1.76 & 1.48 & 89.7x \\
\bottomrule
\end{tabular}
}
\caption{\textbf{Sensitivity to Transaction Costs.} We report Sharpe Ratios under varying execution cost assumptions (0--20 basis points per trade). Despite high turnover ($>35x$), momentum assets (NVDA, TSLA, INTC) retain strong risk-adjusted returns even at 20bps, confirming that the kinematic signal captures substantial price moves that exceed typical friction costs.}
\label{tab:sensitivity}
\end{table}

\subsection{Mechanisms of Alpha Generation}

\begin{figure}[htbp]
    \centering
    \includegraphics[width=1.0\textwidth]{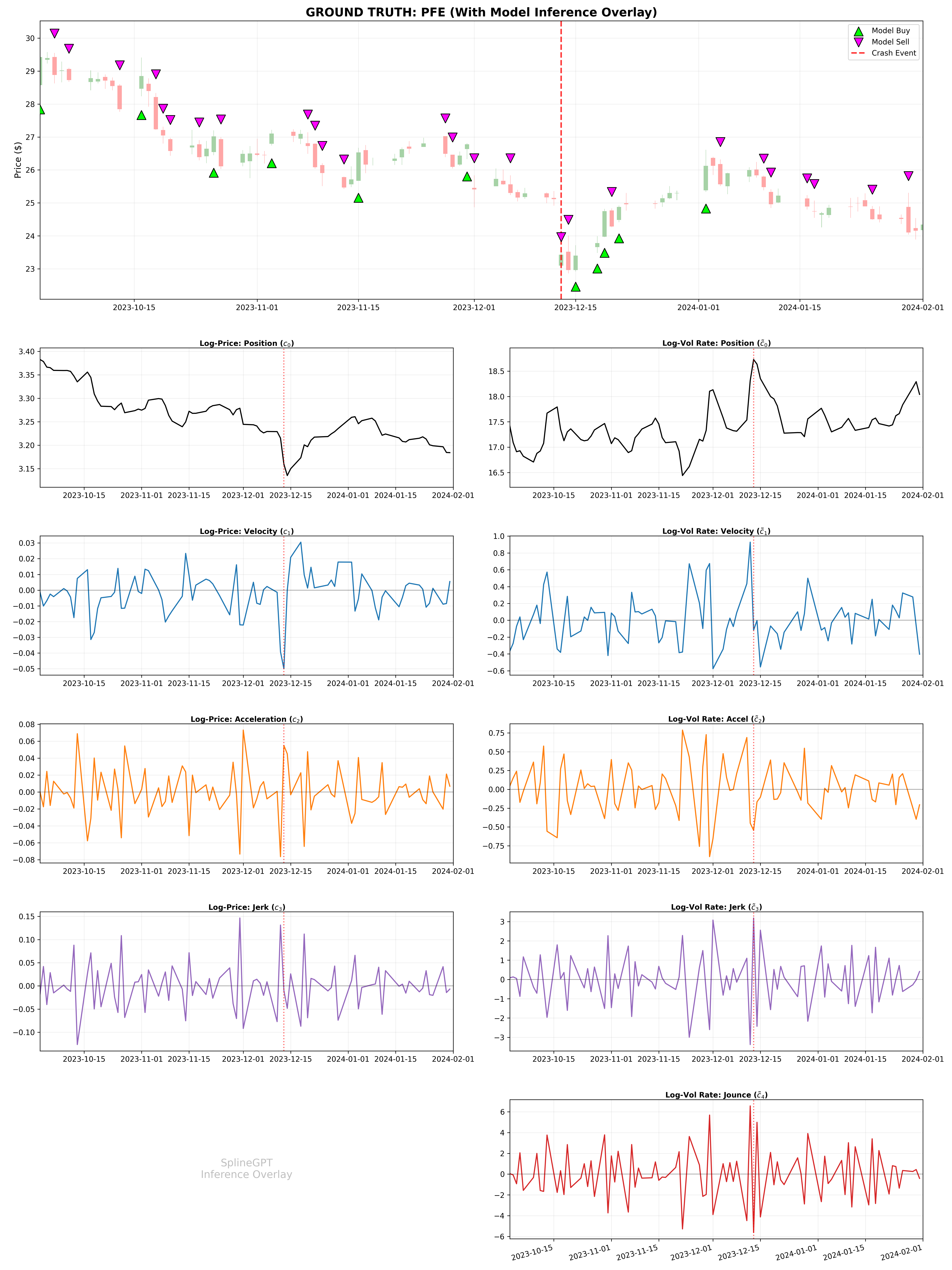}
    \caption{\textbf{The 9-Dimensional Kinematic Manifold vs. Ground Truth.} 
    \textbf{Top Panel:} Standard OHLC Candle and Volume chart for Pfizer (PFE) capturing the capitulation event (Dec 13) and subsequent recovery.
    \textbf{Lower Panels:} The explicit kinematic tokens extracted by \modelname{}. 
    \textbf{Left Column:} Price derivatives ($c_0$--$c_3$). 
    \textbf{Right Column:} Volume derivatives ($\tilde{c}_0$--$\tilde{c}_4$). 
    While the "Ground Truth" (Top) shows a noisy, ambiguous bottoming process, the lower panels reveal the rich, high-frequency signal structure available to the model. The distinct volatility clustering in the higher-order terms (Velocity $c_1$ and Volume Flow Acceleration $\tilde{c}_2$) highlights the complexity of the signal.}
    \label{fig:full_state}
\end{figure}

Given the observed separation between \modelname{} and the baselines exhibiting Liquidation Equilibrium, we highlight two recurring regimes in which the learned policy appears to benefit from smooth coupled derivatives:

\begin{enumerate}
    \item \textbf{Momentum Convexity (TSLA, INTC):} In high-beta assets, price action often exhibits significant convexity (acceleration) during breakouts. \modelname{} successfully captured the positive "Jerk" ($c_3$) associated with the onset of these rallies. By acting on the smooth derivative rather than the noisy price difference, the model entered positions earlier and held through minor volatility that may have induced selling in less sophisticated strategies.
    \item \textbf{Crash Avoidance (PFE):} The most critical validation of the "Active Risk Management" objective is observed in the Pfizer (PFE) case. While the asset shed over 40\% of its value, the model preserved capital (MaxDD -18.8\%). As visualized in Figure \ref{fig:full_state}, the kinematic signatures during a crash are complex, characterized by volatility clustering in the higher-order derivatives. The model's ability to navigate this crash demonstrates its capacity to decouple meaningful structural breaks from transient noise within the 9-dimensional manifold.
\end{enumerate}

% -----------------------------------------------------------
% FIGURE 4: FULL STATE VISUALIZATION
% -----------------------------------------------------------

% -----------------------------------------------------------
% CONCLUSION: GENERALIZABILITY & FUTURE WORK
% -----------------------------------------------------------
\section{Conclusion}
This work presented \modelname{}, a foundation model architecture built upon a novel continuous tokenization pipeline. By treating financial time series as physical observables of a continuous stochastic process, we were able to extract high-order kinematic states (velocity, acceleration, jerk) that are invisible to standard discrete tokenizers.

We validated the model through a rigorous, tax-aware backtest across diverse market sectors. The results demonstrate that the model has learned transferable kinematic regularities that are useful for downstream decision rules, enabling it to:
\begin{enumerate}
    \item \textbf{Kinematic Momentum Capture:} Exploiting positive convexity ($c_2 > 0$) to identify high-beta trend initiations (e.g., TSLA).
    \item \textbf{Structural Break Identification:} Utilizing higher-order derivative decay to detect trend exhaustion prior to price capitulation (e.g., PFE).
    \item \textbf{Cross-Domain Generalization:} Learning robust features that transfer across uncorrelated sectors (Energy, Tech, Pharma) without asset-specific parameter tuning.
\end{enumerate}

Future work will focus on scaling the model parameter count and integrating alternative data modalities, such as macroeconomic indicators, into the continuous token stream. Additionally, while the ``Active Risk Management'' strategy served as a proof of concept, alternative loss weighting schemes could be explored to train adapters for distinct risk profiles. For instance, aggressive ``risk-on'' tunings may be better suited to capture the full upside of parabolic regimes, such as that exhibited by NVDA. Furthermore, the ternary Buy/Sell/Hold classification employed here represents only one potential downstream application. \modelname{} can be fine-tuned for diverse objectives, including regression heads for volatility forecasting or ranking heads for cross-sectional sector rotation.
Future studies will utilize action-rate and calibration diagnostics (e.g., confusion matrices and $\tau$-sweeps) to further disentangle objective-induced abstention from representational insufficiency.

Ultimately, the implications of this work extend beyond finance. We have established a generalized framework for Physics-Informed Tokenization, applicable to any domain where observations are derived from continuous latent states. Unlike finance, when physical first principles exist (e.g. mechanics, electromagnetism, fluid dynamics, etc) the underlying laws are incorporated into the SDE thus modifying the spline structure adaptively for each domain. By formally encoding the governing dynamics into the tokenization layer itself, rather than relying on the model to learn them from scratch, this approach resolves a fundamental dissonance between discrete AI architectures and the continuous physical world. This paradigm shift holds the potential to significantly accelerate the development of robust, physically grounded AI systems.

\printbibliography

@article{Kearney2024,
  title = {Optimization based data enrichment using stochastic dynamical system models},
  author = {Kearney, Griffin M. and Fardad, Makan},
  journal = {PLOS ONE},
  volume = {19},
  number = {9},
  pages = {e0310504},
  year = {2024},
  month = {9},
  publisher = {Public Library of Science},
  doi = {10.1371/journal.pone.0310504},
  url = {https://doi.org/10.1371/journal.pone.0310504}
}

@misc{nie2023timeseriesworth64,
      title={A Time Series is Worth 64 Words: Long-term Forecasting with Transformers}, 
      author={Yuqi Nie and Nam H. Nguyen and Phanwadee Sinthong and Jayant Kalagnanam},
      year={2023},
      eprint={2211.14730},
      archivePrefix={arXiv},
      primaryClass={cs.LG},
      url={https://arxiv.org/abs/2211.14730}, 
}

@misc{liu2024itransformer,
      title={iTransformer: Inverted Transformers Are Effective for Time Series Forecasting}, 
      author={Yong Liu and Tengge Hu and Haoran Zhang and Haixu Wu and Shiyu Wang and Lintao Ma and Mingsheng Long},
      year={2024},
      eprint={2310.06625},
      archivePrefix={arXiv},
      primaryClass={cs.LG},
      url={https://arxiv.org/abs/2310.06625}, 
}

@article{Zhang_2019,
   title={DeepLOB: Deep Convolutional Neural Networks for Limit Order Books},
   volume={67},
   ISSN={1941-0476},
   url={http://dx.doi.org/10.1109/TSP.2019.2907260},
   DOI={10.1109/tsp.2019.2907260},
   number={11},
   journal={IEEE Transactions on Signal Processing},
   publisher={Institute of Electrical and Electronics Engineers (IEEE)},
   author={Zhang, Zihao and Zohren, Stefan and Roberts, Stephen},
   year={2019},
   month=jun, pages={3001–3012} 
}

@article{RAISSI2019686,
title = {Physics-informed neural networks: A deep learning framework for solving forward and inverse problems involving nonlinear partial differential equations},
journal = {Journal of Computational Physics},
volume = {378},
pages = {686-707},
year = {2019},
issn = {0021-9991},
doi = {https://doi.org/10.1016/j.jcp.2018.10.045},
url = {https://www.sciencedirect.com/science/article/pii/S0021999118307125},
author = {M. Raissi and P. Perdikaris and G.E. Karniadakis},
}

@article{Sirignano2019LOB,
author = {Justin A. Sirignano},
title = {Deep learning for limit order books},
journal = {Quantitative Finance},
volume = {19},
number = {4},
pages = {549--570},
year = {2019},
publisher = {Routledge},
doi = {10.1080/14697688.2018.1546053},
}

@misc{shi2024mambastockselectivestatespace,
      title={MambaStock: Selective state space model for stock prediction}, 
      author={Zhuangwei Shi},
      year={2024},
      eprint={2402.18959},
      archivePrefix={arXiv},
      primaryClass={cs.CE},
      url={https://arxiv.org/abs/2402.18959}, 
}

@misc{Ahamed2024TimeMachine,
      title={TimeMachine: A Time Series is Worth 4 Mambas for Long-term Forecasting}, 
      author={Md Atik Ahamed and Qiang Cheng},
      year={2024},
      eprint={2403.09898},
      archivePrefix={arXiv},
      primaryClass={cs.LG},
      url={https://arxiv.org/abs/2403.09898}, 
}

@misc{Hasani2020LTCNN,
      title={Liquid Time-constant Networks}, 
      author={Ramin Hasani and Mathias Lechner and Alexander Amini and Daniela Rus and Radu Grosu},
      year={2020},
      eprint={2006.04439},
      archivePrefix={arXiv},
      primaryClass={cs.LG},
      url={https://arxiv.org/abs/2006.04439}, 
}

@article{Socaciu2025PINN,
  title={Physics-Informed Neural Networks in Pricing Financial Options},
  author={Socaciu, Tibor and Pa{\c{s}}cu, Paul},
  journal={Broad Research in Artificial Intelligence and Neuroscience},
  volume={16},
  number={2},
  pages={474},
  year={2025},
  publisher={EduSoft}
}

@article{liu2025PIKoop,
author = {Liu, Hang and Yan, Gaowei and Cao, Lifeng and Ma, Suxia and Zhao, Guanjia and Liu, Zhongyuan},
title = {Physics-Informed Koopman Networks for Industrial Process Time-Series Prediction},
journal = {Industrial \& Engineering Chemistry Research},
volume = {64},
number = {37},
pages = {18328-18346},
year = {2025},
doi = {10.1021/acs.iecr.5c02217},
}

@inproceedings{Faw2025InContext,
  title={In-Context Fine-Tuning for Time-Series Foundation Models},
  author={Faw, Matthew and Sen, Rajat and Zhou, Yichen and Das, Abhimanyu},
  booktitle={Proceedings of the 42nd International Conference on Machine Learning (ICML)},
  year={2025},
  url={https://icml.cc/virtual/2025/poster/43707}
}

@incollection{LeCun2012,
  author    = {LeCun, Yann A. and Bottou, L{\'e}on and Orr, Genevieve B. and M{\"u}ller, Klaus-Robert},
  title     = {Efficient BackProp},
  booktitle = {Neural Networks: Tricks of the Trade},
  editor    = {Montavon, Gr{\'e}goire and Orr, Genevieve B. and M{\"u}ller, Klaus-Robert},
  series    = {Lecture Notes in Computer Science},
  volume    = {7700},
  pages     = {9--48},
  year      = {2012},
  publisher = {Springer},
  doi       = {10.1007/978-3-642-35289-8_3}
}

@book{Ramsay2005,
  author    = {Ramsay, J. O. and Silverman, B. W.},
  title     = {Functional Data Analysis},
  edition   = {2nd},
  publisher = {Springer},
  address   = {New York},
  year      = {2005},
  doi       = {10.1007/b98888},
}

@book{sarkka2023bayesian,
  title={Bayesian Filtering and Smoothing},
  author={S{\"a}rkk{\"a}, Simo and Svensson, Lennart},
  year={2023},
  edition={2},
  publisher={Cambridge University Press}
}

\end{document}